%% file: root.tex
\documentclass[letterpaper, 10 pt, conference]{ieeeconf}  

\IEEEoverridecommandlockouts                              

\usepackage[letterpaper, margin=0.8in]{geometry}
\usepackage{graphicx}
\usepackage{amsmath}
\usepackage{booktabs}
\usepackage{amssymb}
\usepackage{color}
\title{\LARGE \bf
Latent Disentanglement for Low Light Image Enhancement
}

\author{Zhihao Zheng$^{1}$ , Mooi Choo Chuah$^{1}$
\thanks{*This work was partially supported by National Science Foundation Grant CPS 1931867, and a gift from Qualcomm Technologies, Inc.}
\thanks{$^{1}$Computer Science and Engineering department, P.C. Rossin College of Engineering and Applied Science, Lehigh University, Bethlehem, PA 18015, USA. 
        {\tt\small \{zhzc21@lehigh.edu, chuah@cse.lehigh.edu\}}}%
}

\begin{document}

\maketitle
\thispagestyle{empty}
\pagestyle{empty}

\begin{abstract}

Many learning-based low-light image enhancement (LLIE) algorithms are based on the Retinex theory. However, the Retinex-based decomposition techniques in such models introduce corruptions which limit their enhancement performance. In this paper, we propose a Latent Disentangle-based Enhancement Network (LDE-Net) for low light vision tasks. The latent disentanglement module disentangles the input image in latent space such that no corruption remains in the disentangled Content and Illumination components. For LLIE task, we design a Content-Aware Embedding (CAE) module that utilizes Content features to direct the enhancement of the Illumination component. For downstream tasks (e.g. nighttime UAV tracking and low-light object detection), we develop an effective light-weight enhancer based on the latent disentanglement framework. Comprehensive quantitative and qualitative experiments demonstrate that our LDE-Net significantly outperforms state-of-the-art methods on various LLIE benchmarks. In addition, the great results obtained by applying our framework on the downstream tasks also demonstrate the usefulness of our latent disentanglement design.

\end{abstract}

\input{Body/introduction}
\input{Body/related}

\input{Body/method}

\input{Body/experiments}

\input{Body/conclusions}

\bibliographystyle{IEEEtran}
\bibliography{mybibliography}

\end{document}

%% file: Body/introduction.tex
\section{Introduction}

The emergence of low-cost embedded devices with great computing power and small powerful cameras have
empowered many robotic applications in recent years. For example, one can design visual-based indoor/outdoor navigation schemes for autonomous driving, mobile robots or unmanned aerial vehicles (UAVs) or visual-based object tracking. 
Deep learning based models have been designed for such robotic applications. Unfortunately, such models are typically trained using images captured during day time. In real world deployments, we typically encounter scenarios with varying light conditions e.g. night-time operations. Images captured under low-light conditions typically contain noise, low contrasts and result in performance degradation using models trained on normal light images. To overcome such limitations, much research has been spent on designing low-light image enhancement techniques.

In the past, several traditional adjustment techniques based on the domain knowledge and statistical properties (e.g., Gamma correction and histogram equalization) have been widely employed for the Low-Light Image Enhancement (LLIE) task and showed satisfactory results for globally-degraded low-light images. However, such traditional approaches usually suffer from poor generalization and robustness on images captured under varying light conditions. 

Many deep learning-based LLIE methods have been proposed to further address these drawbacks. Some general methods regard low-light image enhancement as a restoration task and construct models to learn an overall mapping between low-light and normal-light images. Although such general methods achieve better results than traditional approaches, 
most of them still exhibit unsatisfactory enhancement results such as uneven illumination, low efficiency, lack of robustness to noise and structure details etc. 

Decomposition-based LLIE methods first decompose images into two components and then design specific restoration and enhancement technique for each component. Among all these decomposition methods, LLIE models based on Retinex theory have attracted the most attention. Retinex LLIE methods usually decompose images into a illumination map and a reflectance image, then learn to bright up the illumination map and subsequently fuse the improved illumination map with the denoised reflectance image. They impose a structure-aware loss function to constrain the enhancement of the illumination component during training but such design results in poorer decomposition results.

Based on the above observation, we propose a latent disentanglement framework for low-light image enhancement and other downstream vision tasks (e.g. object detection and tracking). Our design disentangles images into light-specific component (i.e. Illumination) and light-invariant component (i.e. Content) in latent feature space for better disentangle results. Then, only the Illumination component requires enhancement during the enhancement stage since the features for the Content component are invariant under varying light conditions. Furthermore, we propose a Content-Aware Embedding module (CAE) to  explicitly learn the correlation between the Content and Illumination features, thereby improving the performance of the Illumination enhancement network.
Last but not least, taking advantage of our proposed latent disentanglement design, we can easily develop very light-weight enhancement networks to improve performance for downstream low light vision tasks (e.g. nighttime UAV tracking and low-light object detection). 

\begin{figure*}[!t]
    \centering
        \includegraphics[width=0.9\textwidth]{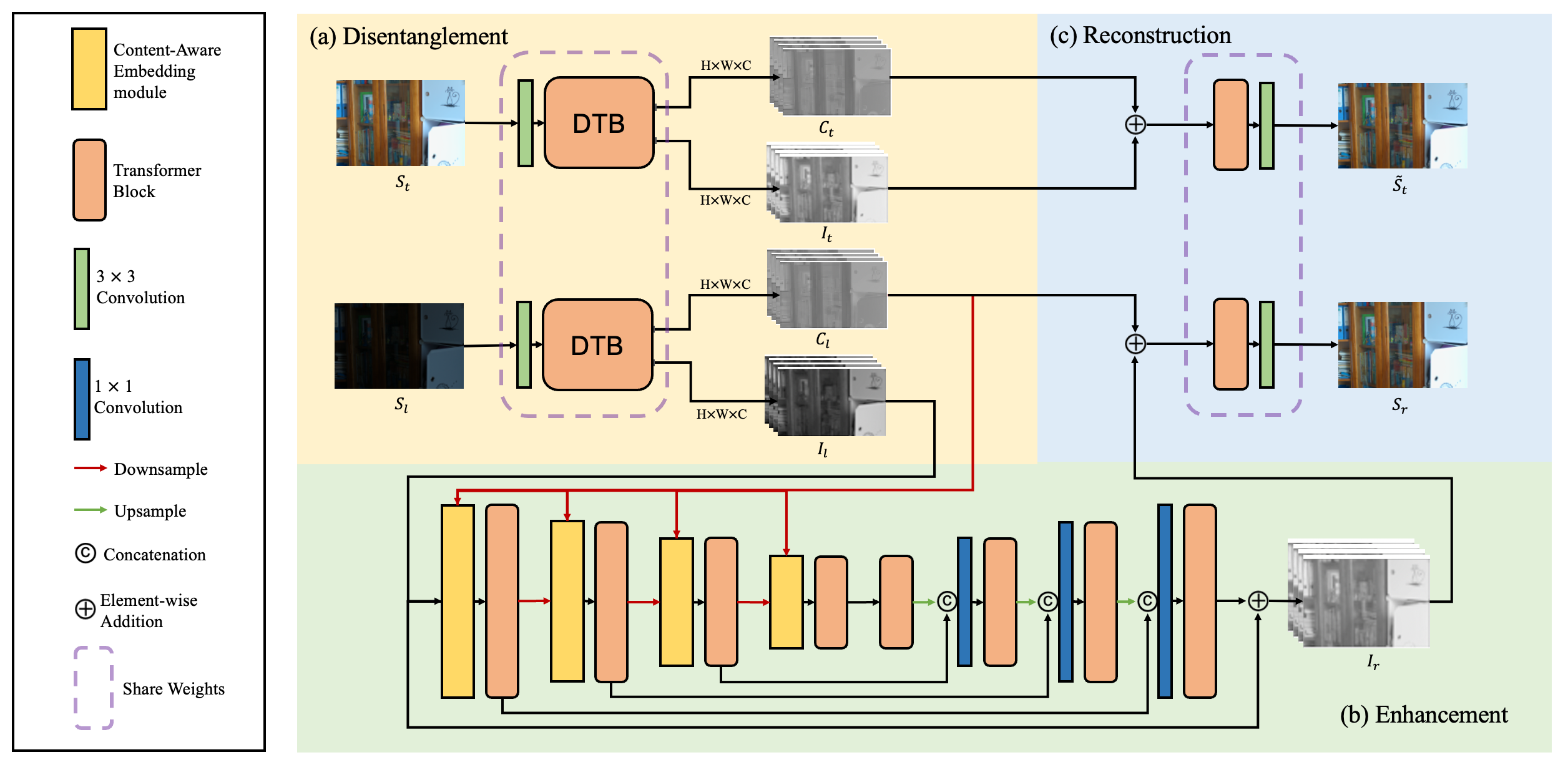}
    \caption{Overview of the proposed LDE-Net. (a)The disentanglement module disentangle an input image to the Content and Illumination components, both of which are latent space features. (b)Content-Aware Illumination Enhancement module enhances the Illumination with the guidance of Content feature. (c)The reconstruction module reconstructs a new image with the Content and restored Illumination component.}
    \label{Fig:framework}
    \vspace{-4mm}
\end{figure*}

In summary, the contributions of this paper are summarized as:
\begin{itemize}
    \item Unlike existing Retinex-based decomposition methods, we propose a transformer-based latent disentanglement framework to get better disentanglement results for low-light image enhancement and other downstream low-light vision tasks. 
    \item We design a Content-Aware Embedding module (CAE), which explicitly learns the correlation between the disentangled Content and Illumination features, thereby improving the performance of our Latent Disentangle-based Enhancement Network (LDE-Net). Extensive experimental results using three public benchmark datasets demonstrate that our proposed enhancement network outperforms SOTA methods.
    \item Benefiting from our proposed latent disentanglement design, we implement extremely light weight enhancement network for downstream low light vision tasks (e.g. nighttime UAV tracking and low-light object detection) that boost the performance of existing methods with proper task-tailored training. 
\end{itemize}

%% file: Body/related.tex
\section{Related Work}
\label{sec:related}

\subsection{Traditional LLIE methods}

Traditional methods for low-light image enhancement include Histogram Equalization-based methods \cite{abdullah2007dynamic} and Retinex-based methods \cite{jobson1997properties}. The former enhances low-light images by extending the dynamic range. The latter first decomposes a low-light image into a reflectance image and an illumination map, and the reflectance image is further improved to generate the enhanced image.  Such traditional methods rely on hand-crafted priors, but designing proper priors for various light conditions is difficult.

\subsection{Learning-based LLIE methods}. 

Recent deep learning based methods show promising results \cite{jiang2022degrade, wang2020deep, afifi2020deep, wei2018deep}. We can further divide existing designs into Retinex-based methods and end-to-end methods. 

\textbf{Retinex-based methods.} Retinex-based methods use deep network to decompose and enhance an image. Wei et al. proposed a two-stage Retinex-based method called Retinex-Net \cite{wei2018deep}. Inspired by Retinex-Net, Zhang et al. proposed two refined methods, called KinD \cite{zhang2019kindling} and KinD++ \cite{zhang2021beyond}. Recently, Wu et al. proposed a novel deep unfolding Retinex-based network to further integrate the strengths of model-based and learning-based methods \cite{wu2022uretinex}.

\textbf{End-to-End methods.}  In comparison to Retinex-based method, recent end-to-end methods directly learn an overall mapping to generate an enhanced image \cite{dong2022abandoning, dudhane2022burst, fan2022half, tu2022maxim, wang2022low, xu2020learning, xu2022snr, zheng2023robustness, yao2023goal,  zheng2021adaptive, zhu2020eemefn}. Lore et al. \cite{lore2017llnet} made the first attempt by proposing a deep autoencoder model named Low-Light Net (LLNet). Later on, various end-to-end methods are proposed. Physics-based concepts, e.g. Laplacian pyramid \cite{lim2020dslr}, local parametric filter \cite{moran2020deeplpf}, De-Bayer-Filter \cite{dong2022abandoning}, normalization flow \cite{wang2022low} and wavelet transform \cite{fan2022half}, are proposed to improve model interpretability and lead to more satisfactory results. In \cite{jiang2021enlightengan, jin2022unsupervised}, adversarial learning is introduced to capture the visual properties. In \cite{guo2020zero}, \cite{cui2022you}, the light enhancement is creatively formulated as a task of image specific curve estimation using zero-shot learning. In \cite{kim2022learning, yang2022adaint}, 3D lookup table and color histogram are utilized to preserve the color consistency. Recently, Transformer-based methods have been applied to LLIE tasks. Wang et al. \cite{wang2022uformer} adopts a modified swin-transformer block to build a U-shaped network; Zamir et al. \cite{zamir2022restormer} introduces a transposed transformer block to improve feature aggregation for image restoration; Wang et al. \cite{wang2023ultra} design an Axis-based transformer block to further improve efficiency of the attention mechanism on LLIE tasks.

%% file: Body/method.tex
\section{Methodologies}
\label{sec:method}
\subsection{Motivation}


It is extremely challenging to extract the illumination component from a single image.
Most of the previous decomposition models are based on the well-known Retinex theory. According to the Retinex theory, an input image $S \in R^{h \times w \times 3}$ can be decomposed into a 3-channel reflectance image $R \in R^{h \times w \times 3}$and a one-channel illumination map $I \in R^{h \times w \times 1}$ as 

\begin{equation}
 S = R \odot I
\end{equation}

where, $\odot$ denotes the element-wise multiplication. 
Modern Retinex methods use deep learning models to estimate the illumination map $I$ and corresponding reflectance image $R $ using the above equation. Small errors randomly generated by such models will be amplified by this dot-product related operation and lead to inaccurate estimation. Thus, previous Retinex-based decomposition models \cite{wei2018deep}, \cite{zhang2020self}, \cite{zhang2019kindling}, \cite{wu2022uretinex}, \cite{yi2023diff} cannot obtain satisfactory decomposition results and hence require further enhancement to the decomposed reflectance images during the adjustment step. To overcome this problem, we use a disentanglement strategy inspired by Causal theory to extract the illumination component from the input image by operating in the latent feature space instead of the RGB image pixel space as in the traditional Retinex-based approaches. 

\begin{figure}[t]
        \includegraphics[width=0.5\textwidth]{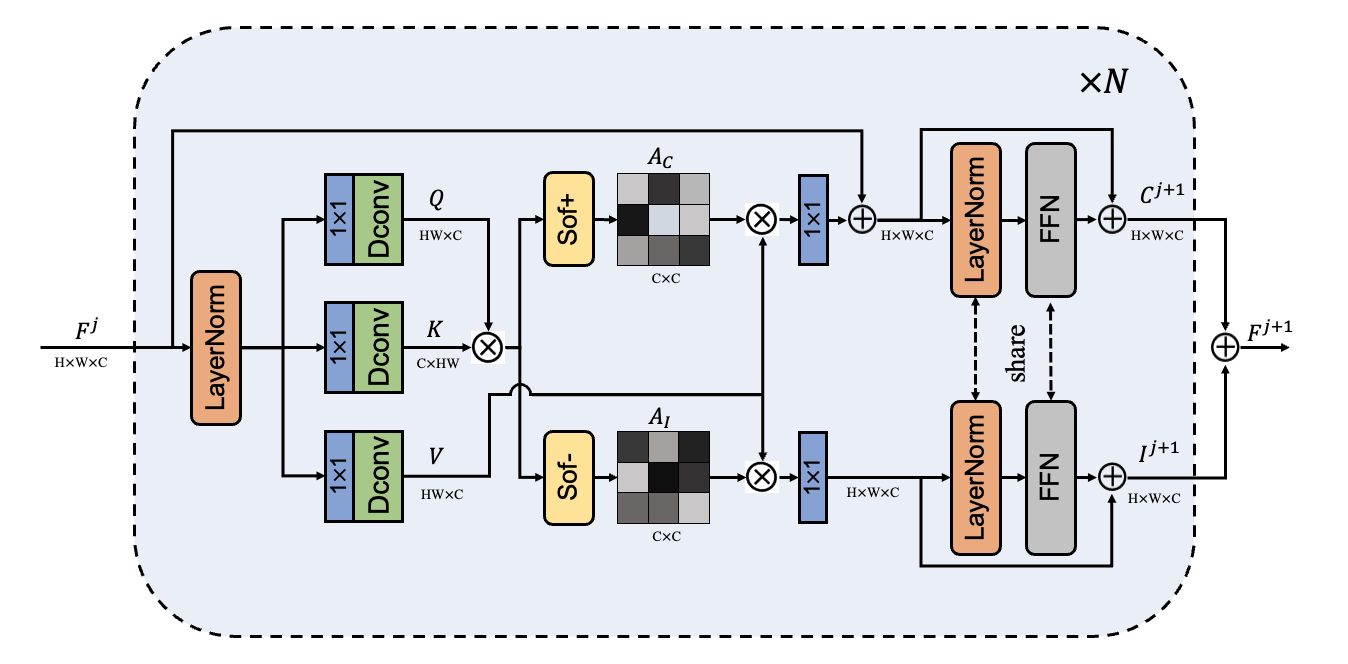}
    \caption{Illustration of the implementation details of the Disentanglement Transformer Block. (DTB)}
    \label{Fig:DAM}
    \vspace{-4mm}
\end{figure}

\subsection{Overview}
In this work, we propose a Latent Disentangle-based Enhancement Network (LDE-Net) as illustrated in Fig \ref{Fig:framework} which consists of three modules: the disentanglement module, the enhancement module and the reconstruction module.  The overall model operations can be represented as follows:

\begin{equation}
 S_r = M_{recon}(M_{enh}(I, C), C) = M_{recon}(M_{enh}(M_{dis}(S)))
\end{equation}

where $S$ is an input image and $S_r$ is the restored image. The disentanglement module $M_{dis}$ disentangle images into the light-invariant Content $C$ and the light-specific Illumination component $I$ , both of which are latent space features, as shown in Fig \ref{Fig:framework} (a). Then, the enhancement module $M_{enh}$ incorporates both Content and Illumination component to bright up the illumination,  as illustrated in Fig \ref{Fig:framework} (b). Finally, the reconstruction module $M_{recon}$ reconstructs a new image with the Content and restored Illumination component
, as shown in Fig \ref{Fig:framework} (c).

\subsection{Latent Disentanglement}
\label{sec: latent dis}

Different from previous decomposition models, our design is a disentangle-then-reconstruct architecture.  To make disentanglement light weight and preserve high-resolution information, we avoid using hierarchical structure for both disentanglement and reconstruction modules. Given an input image $\mathbf{S} \in R^{h \times w \times 3}$, the disentanglement module $M_{dis}$  first employs a $3 \times 3$ convolution as a embedding layer to extract shallow feature $\mathbf{F}_0 \in R^{h \times w \times c} $. Next, the Disentangle Transformer Blocks (DTB) disentangle $\mathbf{F}_0$ into the Content component $\mathbf{C} \in R^{h \times w \times c} $ and Illumination component $\mathbf{I} \in R^{h \times w \times c} $ in latent space. The disentanglement module $M_{dis}$ can be formulated as:

\begin{equation}
\mathbf{C} , \mathbf{I} = M_{dis}(\mathbf{S})
\end{equation}

Then, the Content $\mathbf{C}$ and Illumination $\mathbf{I}$ are combined together as the input to the following transposed transformer blocks \cite{zamir2022restormer} in the reconstruction module $M_{recon}$. Finally,  we apply a  $3 \times 3$ convolution to yield a new image $\tilde{\mathbf{S}}\in R^{h \times w \times 3}$. The reconstruction module $M_{recon}$ can be described as:

\begin{equation}
 \tilde{\mathbf{S}} =  M_{recon}(\mathbf{C}+\mathbf{I})
\end{equation}

\begin{table*}[!t]
    
    \tiny
    \begin{center}
    \resizebox{0.85\textwidth}{!}{
    \begin{tabular}{c|cc|cc|cc}
    
        \toprule
           &  \multicolumn{2}{c|}{LOL-v1 Dataset} &  \multicolumn{2}{c|}{LOL-v2 Dataset}  &  \multicolumn{2}{c}{LDIS Dataset} \\
        \hline
             Methods  & PSNR$(\uparrow)$ & SSIM$(\uparrow) $ & PSNR$(\uparrow)$ & SSIM$(\uparrow)$ & PSNR$(\uparrow)$ & SSIM$(\uparrow)$\\
        \hline
           Retinex-Net \cite{wei2018deep} & 16.77 & 0.560 & 15.47 & 0.567 &16.12 & 0.573  \\

            KinD++ \cite{zhang2019kindling} & 18.45 & 0.779  & 18.63 & 0.799 & 16.52 & 0.582 \\
            
           LIME \cite{guo2016lime} & 16.05 & 0.486 &17.16 & 0.480 & 16.25 & 0.539 \\
       
           IAT \cite{cui2022you} &  23.38  & 0.809 & \textcolor{blue}{23.50} & 0.824  &17.34 & 0.692 \\

            LLFlow \cite{wang2022low}  & 21.15 & \textcolor{blue}{0.852} & 22.37 & \textcolor{blue}{0.865}  & 18.01 &  \textcolor{blue}{0.789 } \\
            
           Restormer \cite{zamir2022restormer} & 22.37  & 0.816 & 19.94 & 0.827  &17.13 &  0.727 \\


            Uformer \cite{wang2022uformer} & 19.61 & 0.755 & 19.41 & 0.657 &  17.01 &  0.694 \\

           LLFormer \cite{wang2023ultra} & \textcolor{blue}{23.65}  & 0.816 & 21.46 & 0.821  & \textcolor{blue}{18.67} &  0.685 \\

           LDE-Net(our)  &  \textcolor{red}{25.02} & \textcolor{red}{0.910} & \textcolor{red}{25.49} & \textcolor{red}{0.950}  & \textcolor{red}{19.38} &  \textcolor{red}{0.848}  \\
         \bottomrule
    \end{tabular}
    }
    \end{center}
    \caption{ Quantitative comparisons on LOL-v1, LOL-v2 and LDIS datasets. The highest result is in \textcolor{red}{red} color while the second highest result is in \textcolor{blue}{blue} color. Our LDE-Net significantly outperforms SOTA algorithms.}
    \vspace{-4mm}
    \label{tab:results}
\end{table*}

\begin{figure}[t]
        \includegraphics[width=0.5\textwidth]{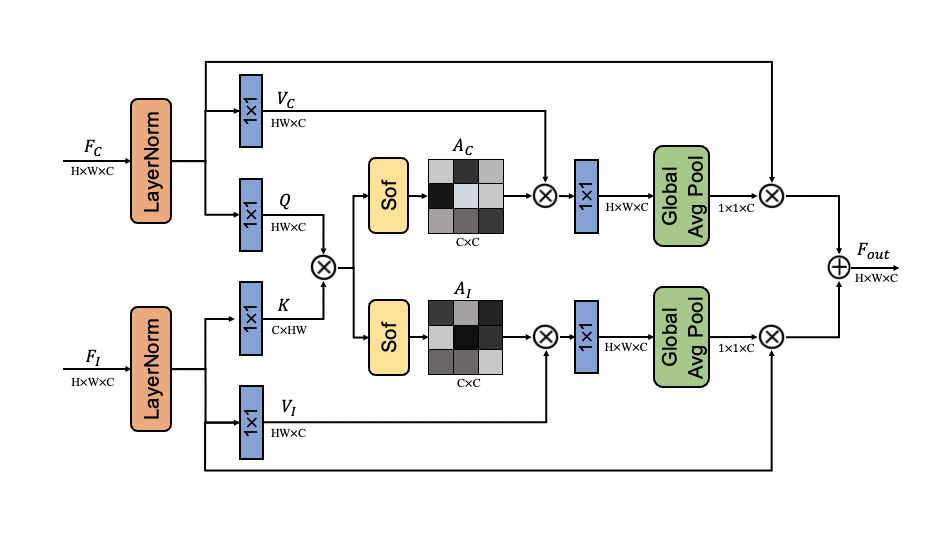}
    \caption{Illustration of the implementation details of the Content-Aware Embedding Module (CAE). }
    \label{Fig:CAE}
    \vspace{-4mm}
\end{figure}

\textbf{Disentanglement Transformer Block.} We build Disentanglement Transformer Block (DTB) based on the transposed transformer block in Restormer \cite{zamir2022restormer} where the self-attention (SA) mechanism is applied across feature dimension rather than the spatial dimension. 

As shown in Fig \ref{Fig:DAM}, given a feature $\mathbf{Y}\in  R^{h \times w \times c}$ output from the normalization layer, we first generate query ($\mathbf{Q}$), key ($\mathbf{K}$) and value ($\mathbf{V}$) projections by applying $1 \times 1$ convolutions followed by $3\times3$ depth-wise convolutions to encode channel-wise spatial context, as $\mathbf{Q} = W^Q_d W^Q_p \mathbf{Y}$, $\mathbf{K} = W^K_d W^K_p \mathbf{Y}$, $\mathbf{V} = W^V_d W^V_p \mathbf{Y}$, where $W_p^{(\cdot)}$ is the $1 \times 1$ point-wise convolution and $W_d^{(\cdot)}$ is the $3\times3$ depth-wise convolution. Next, we reshape query and key so that their dot-product generates a transposed-attention map $\mathbf{A}\in  R^{c  \times c}$. Then, we calculate Content attention $ \mathbf{A}_C$ and Illumination attention $\mathbf{A}_I$  through two Softmax functions. Overall, the disentanglement attentions are formulated as:




\begin{equation}
    \begin{aligned}
   & \mathbf{Q}, \mathbf{K}, \mathbf{V} = W^Q_d W^Q_p \mathbf{Y}, W^K_d W^K_p \mathbf{Y}, W^V_d W^V_p\mathbf{Y} \\
  &  \mathbf{A}_C = Softmax(\hat{\mathbf{Q}} \cdot \hat{\mathbf{K}^T}) \mathbf{V}  \\
   & \mathbf{A}_I = Softmax(-\hat{\mathbf{Q}} \cdot \hat{\mathbf{K}^T}) \mathbf{V}  
    \end{aligned}
\end{equation}

where, $\hat{\mathbf{Q}} \in R^{hw \times c} $,  $\hat{\mathbf{K}} \in R^{hw \times c} $ and  $\hat{\mathbf{V}} \in R^{hw \times c} $ are obtained after reshaping tensors from the original size $R^{h \times w \times c}$. 

Similar to the conventional vision transformer \cite{yuan2021tokens}, both Content attention $ \mathbf{A}_C$ and Illumination attention $\mathbf{A}_I$ are fed into a normalization layer and Feed-Forward Network(FFN) to generate the Content and Illumination features. To get better disentanglement results, we implement $N$  Disentanglement Transformer Blocks(DTB) in the disentanglement module (In this work, we set $N=2$). Given the input feature $\mathbf{F}^j$, the ${j+1}^{th}$ DTB can be formulated as follows:

\begin{equation}
    \begin{aligned}
    &\mathbf{A}_C^{j+1}, \mathbf{A}_I^{j+1}  = DAM(LN(\mathbf{F}^j)) \\
   &\mathbf{C}^{j+1} = FFN(LN(\mathbf{A}_C^{j+1} + \mathbf{F}^j)) + \mathbf{A}_C^{j+1} + \mathbf{F}^j  \\
   & \mathbf{I}^{j+1} = FFN(LN(\mathbf{A}_I^{j+1})) + \mathbf{A}_I^{j+1} \\ 
   & \hat{\mathbf{F}}^{j+1} = \mathbf{C}^{j+1} + \mathbf{I}^{j+1}
    \end{aligned}
\end{equation}

where, $\mathbf{F}^{j+1}$ is the output feature, $LN(\cdot)$ represents the normalization layer. Notice that the Content feature $\mathbf{C}^N$ and Illumination feature $\mathbf{I}^N$ in the last block $N$ are the final output from the distanglement module. 

\begin{figure*}[t]
    \centering
        \includegraphics[width=\textwidth]{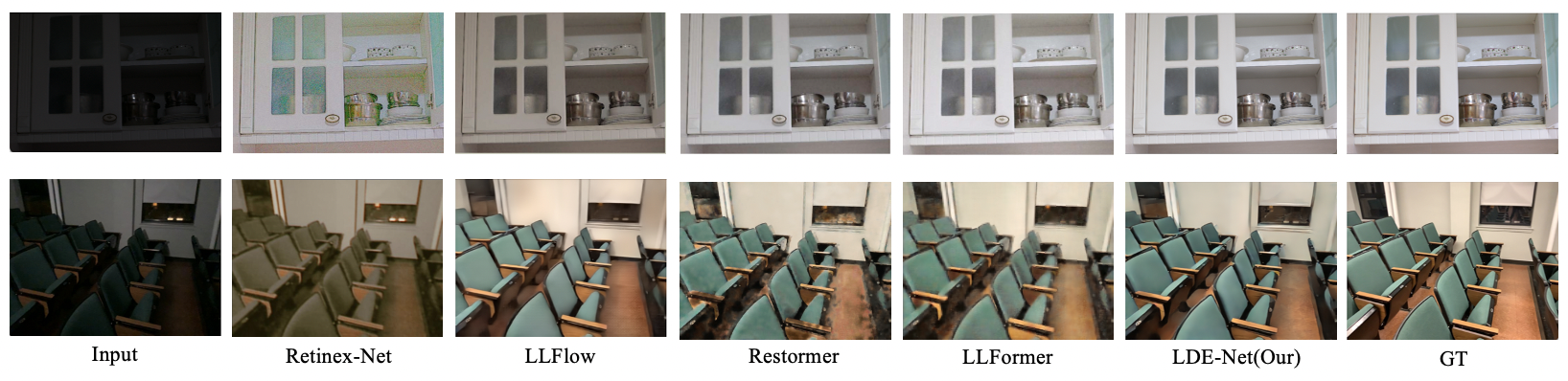}
    \caption{Qualitative results of LOL (top) and LDIS (bottom). Previous methods either collapse by noise, or distort color, or produce blurry and under-/over-exposed images. While our algorithm can effectively reconstruct well-exposed image details.}
    \label{Fig:qualitative}
    \vspace{-4mm}
\end{figure*}

\textbf{Disentanglement Loss Functions.}  Since we do not have the ground-truth disentanglement of real images, we follow the training approach proposed in \cite{wei2018deep}. The disentanglement module takes the paired low-light image $\mathbf{S}_l$ and the normal-light image $\mathbf{S}_t$ as inputs, then estimates the Content $\mathbf{C}_l$ and the Illumination $\mathbf{I}_l$ for $\mathbf{S}_l$, $\mathbf{C}_t$ and $\mathbf{I}_t$ for $\mathbf{S}_t$, respectively.

\begin{equation}
    \begin{aligned}
    &\mathbf{C}_l, \mathbf{I}_l  = M_{dis}(\mathbf{S}_l) \\
   &\mathbf{C}_t, \mathbf{I}_t  = M_{dis}(\mathbf{S}_t)
    \end{aligned}
    \label{Eq: dis}
\end{equation}

Recall that the Content component is corruption-free and invariant under different light condition, we impose strong constraints between low-light and normal-light Content components. The Content consistency loss $L_{cc}$ is introduced to constrain the consistency of Content components:

\begin{equation}
    L_{cc} = \Vert \mathbf{C}_l - \mathbf{C}_t \Vert_1
\end{equation}

Based on our proposed latent disentanglement design, we employ no constraints on Illumination component. Then, we apply the reconstruction loss $L_{recon}$ to ensure that 
both $\mathbf{C}_l$ and $\mathbf{C}_t$ can reconstruct a new image that is similar to the input image $\mathbf{S}_l / \mathbf{S}_t$ with the corresponding Illumination component $\mathbf{I}_l / \mathbf{I}_t$. The reconstruction loss $L_{recon}$ can be formally described as:

\begin{equation}
    L_{recon} = \sum_{i=l}^{t} \sum_{j=l}^{t} \Vert M_{recon}(\mathbf{C}_i+\mathbf{I}_j) - \mathbf{S}_j \Vert_1
    \label{Eq: loss recon}
\end{equation}

Overall, the disentanglement loss $L_{dis}$ can be formulated as:

\begin{equation}
    L_{dis} = L_{cc} + \lambda_{recon} L_{recon}
\label{Eq: loss dis}
\end{equation}

where $\lambda_{recon}$ denote the coefficient to balance the two losses.

\begin{figure}[t]
        \includegraphics[width=0.5\textwidth]{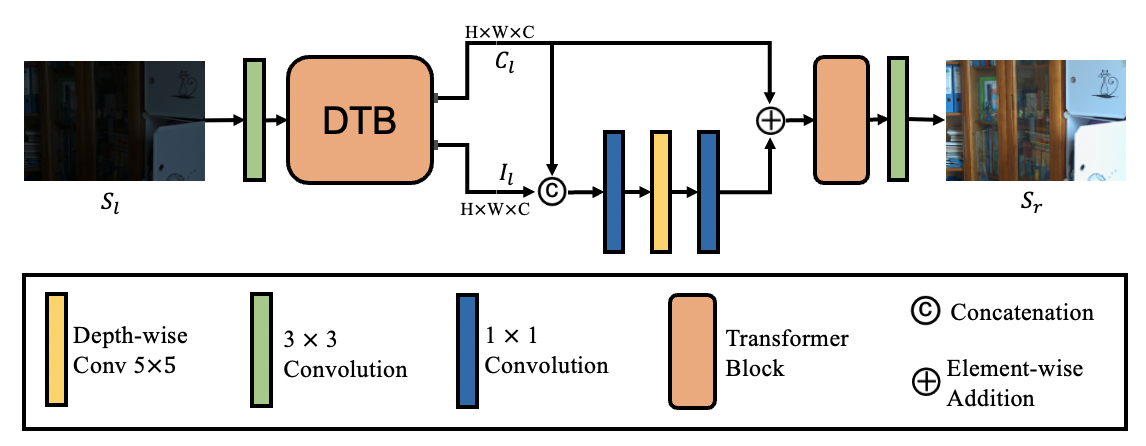}
    \caption{Overview of the light-weight enhancer with latent disentanglement module. }
    \label{Fig:light-enh}
    \vspace{-4mm}
\end{figure}

\subsection{Content-Aware Illumination Enhancement}

The output of the disentanglement module yields both the Content and Illumination features. We only need to restore the degraded low-light Illumination $\mathbf{I}_l$ to normal-light $\mathbf{I}_t$.

As in previous restoration work \cite{zamir2022restormer}, \cite{wang2023ultra}, \cite{wang2022uformer}, our illumination enhancement module is a Transformer-based 4-level hierarchical encoder-decoder network,  as shown in Fig. \ref{Fig:framework} (b). From top to bottom levels, the encoder-decoder hierarchically reduces spatial dimension and doubles the channel dimension. We use our proposed Content-Aware Embedding (CAE) module to adaptively fuse the Illumination feature with the Content feature at the input of each encoder block. For feature downsampling and upsampling, we apply pixel-unshuffle and pixel-shuffle operations \cite{shi2016real}. To help recover high-resolution representations, the encoder features are concatenated with the decoder features via skip connections \cite{ronneberger2015u}.

\textbf{Content-Aware Embedding Module.}
When restoring Illumination component with the guidance of Content component, the main challenge is to efficiently exchange information between the two components and adaptively fuse them together. To tackle this issue, we propose Content-Aware Embedding (CAE) module to refine the Illumination features as shown in Fig.\ref{Fig:CAE}. Our CAE design uses the attention mechanism so as to align with our Transformer backbone. 

Our CAE module is designed based on the Cross-Attention mechanism that takes both source feature (e.g. Content feature $\mathbf{F}_C$) and target feature (e.g. Illumination feature $\mathbf{F}_I$) as inputs and fuse the source feature with target feature. In order to fuse high-resolution features with a low computational cost, we use transposed-attention mechanism \cite{zamir2022restormer} to calculate channel-wised attentions with the input features. Moreover, we use the Global Average Pooling function to learn adaptive fuse weights instead of directly fuse dense features from the cross-attention layer to reduce the training difficulty. 




\subsection{Loss functions for LLIE}

The training objective for the LLIE task is to minimize the difference between the normal-light image and enhanced image. Thus, our loss functions for LLIE as follows:

\begin{equation}
     \begin{aligned}
        L_{enh} = \Vert \mathbf{S}_r - \mathbf{S}_t \Vert_1 + & \lambda_{s}(1-SSIM(\mathbf{S}_r, \mathbf{S}_t))   \\
         & + \lambda_{p} \Vert \theta(\mathbf{S}_r) - \theta(\mathbf{S}_t) \Vert_2  
     \end{aligned}
    \label{Eq: loss enh}
\end{equation}

where, $\mathbf{S}_r$ and  $\mathbf{S}_t$ represent enhanced image and corresponding normal-light image. $SSIM( \cdot, \cdot)$ is structural similarity \cite{wang2004image} and $\theta(\cdot)$ denotes the process to extract deep features form a pre-trained network. $\lambda_s$ and $\lambda_p$ are the weighting factors for balancing the relative importance of each item.

%% file: Body/experiments.tex
\section{Experiments}

\subsection{Experimental Settings}

\textbf{Datasets.} We evaluate the proposed LDENet on three datasets involving both indoor and outdoor scenes, namely LOL-v1\cite{wei2018deep}, LOL-v2\cite{yang2021sparse} and LDIS \cite{ying2022delving}. The LOL-v1 dataset \cite{wei2018deep} is a real captured dataset including 485 low/normal light image pairs for training and 15 pairs for testing. The LOL-v2 dataset contains 689 image pairs for training and 100 pairs for testing. The low-light images of LOL-v1/v2 are collected by changing exposure time and ISO of normal light images.  The Light-Dark Indoor Scenes (LDIS) dataset \cite{ying2022delving} consists of 487 low/normal-light image pairs collected from 87 different indoor scenes under both well-illuminated and low-light conditions. Although LDIS is collected for low light semantic segmentation tasks, it is collected using a mobile device (e.g. an Apple iPad Pro $4^{th}$ Gen) under real world condition (e.g. by turning on/off all the lights). Compared with previous datasets, such realistic low light dataset offers more diverse lighting conditions for the enhancement task.

\textbf{Metrics.} We adopt the widely used peak signal-to-noise ratio (PSNR) and structural similarity index (SSIM) \cite{wang2004image}as the evaluation measures to compare our proposed method with others.

\textbf{Implementation Details.}  We implement the proposed framework using PyTorch. The model is trained with the Adam \cite{kingma2014adam} optimizer ($\beta_1$ = 0.9 and $\beta_2$ = 0.999). The learning rate is initially set to $ 2 \times 10^{-4}$ and then steadily decreased to $ 1 \times 10^{-6} $ by the cosine annealing scheme \cite{loshchilov2016sgdr} during the training process. Patches at the size of  $128 \times 128$ are randomly cropped from the low/normal-light image pairs as training samples. The batch size is 8. The training data is augmented with random HSV, rotation and flipping. Our disentanglement module and enhancement module are trained with the loss function as shown in Eq. \ref{Eq: loss dis} and Eq. \ref{Eq: loss enh}, and we empirically set the parameter $\lambda_{recon} = 0.2$, $\lambda_p = \lambda_s = 1$.

\begin{table*}[t]
    \small
    \renewcommand{\arraystretch}{1.2}
    \begin{center}
    \resizebox{\textwidth}{!}{
    \begin{tabular}{c|ccc|ccc|ccc|ccc}
        \toprule
        Tracker & \multicolumn{3}{c|}{SiamAPN++ \cite{cao2021siamapn++}} & \multicolumn{3}{c|}{SiamRPN++ \cite{li2019siamrpn++}} & \multicolumn{3}{c|}{DiMP18 \cite{bhat2019learning}} & \multicolumn{3}{c}{DiMP50 \cite{bhat2019learning}} \\
         \hline
          Enhancer & Baseline & w. SCT & w. Ours  & Baseline & w. SCT &  w. Ours 
         & Baseline & w. SCT & w. Ours & Baseline &  w. SCT &  w. Ours\\
        \hline
           Success Rate & 0.375 & 0.406 & 0.411 & 0.372 & 0.432 & 0.454 & 0.468 & 0.517 & 0.518 & 0.495 & 0.518 & 0.524 \\

           Precision & 0.494 & 0.542 & 0.551 & 0.515 & 0.592 & 0.595 & 0.624 & 0.678 & 0.678 & 0.644 & 0.681 & 0.679 \\

           FPS & 194.9 & 87.2 & 81.6 & 119.3 & 52.5 & 54.9 & 52.9 & 40.3 & 37.1 & 35.4 & 29.3 & 29.0 \\
         \bottomrule
    \end{tabular}
    }
    \end{center}
    \caption{Nighttime tracking performance on DarkTrack2021 dataset. Baseline: the original tracker without any enhancer. w.SCT: with SCT activated, w.Ours: with our light-weight enhancer activated}
    \vspace{-4mm}
    \label{tab:tracking results}
\end{table*}

\begin{table}[t]
    \small
    \renewcommand{\arraystretch}{1.4}
    \begin{center}
    \resizebox{0.5\textwidth}{!}{
    \begin{tabular}{c|cc|cc}
        \toprule
           &  \multicolumn{2}{c|}{LOL-v1}   &  \multicolumn{2}{c}{LDIS} \\
        \hline
             Metrics  & Retinex-Net & Ours  & Retinex-Net  & Ours  \\
        \hline
          $PSNR(M_{r}(\mathbf{C}_l,\mathbf{I}_l) , \mathbf{S}_l)(\uparrow)$ & 44.81 &  48.32 & 51.81 & 48.34 \\
        \hline
          $PSNR(M_{r}(\mathbf{C}_t,\mathbf{I}_t) , \mathbf{S}_t)(\uparrow)$ & 35.84 &48.46 & 43.94 & 50.38  \\
        \hline
          $PSNR(M_{r}(\mathbf{C}_t,\mathbf{I}_l) , \mathbf{S}_l)(\uparrow)$ & 37.58 &36.85 & 30.25 & 37.22 \\
        \hline
          $PSNR(M_{r}(\mathbf{C}_l,\mathbf{I}_t) , \mathbf{S}_t)(\uparrow)$ & 22.80 &34.92  & 17.19 & 36.02  \\
         \bottomrule
    \end{tabular}
    }
    \end{center}
    \caption{We conduct cross-dataset ablation study to the proposed latent disentanglement. We pre-train on LOL-v1 dataset and test on LOL-v1, LOL-v2 and LDIS dataset.}
    \vspace{-4mm}
    \label{tab:cross-dataset dis}
\end{table}

\subsection{Low-light Image Enhancement}

\textbf{Quantitative Results.} We quantitatively compare the proposed method with conventional  and SOTA enhancement algorithms in Table \ref{tab:results}. The evaluation code for KinD++ \cite{zhang2019kindling} and LLFlow \cite{wang2022low} models involve an unreasonable adjustment to the enhanced images using ground-truth images so we evaluated their models using the corrected evaluation code. As shown in Table~ \ref{tab:results}, our method outperforms the best SOTA model on three datasets in all metrics. Compared with the LLFormer \cite{wang2023ultra} and LLFlow \cite{wang2022low} , our method improves PSNR by 1.73 dB and significantly increases SSIM by 5.8 \% on LOL-v1 datasets.  Our LDE-Net also outperforms IAT \cite{cui2022you} and LLFlow \cite{wang2022low}  on the LOL-v2 dataset, achieving a significant gain of PSNR by 1.99 dB and SSIM by 8.5 \%. This trend is also observed on the more challenging LDIS dataset as shown in Table~\ref{tab:results}. Our approach consistently outperforms other methods in all metrics. All these results clearly suggest the effectiveness of our LDE-Net, especially on more challenging and diverse conditions.

\textbf{Qualitative Results.} The visual comparisons of the proposed LDE-Net and SOTA algorithms are shown in Fig. \ref{Fig:qualitative}. Please zoom in for a better review. The top row of Fig. \ref{Fig:qualitative} shows the original image from LOL-v1 and its corresponding enhanced images using different models. One can see that Retinex-Net causes serious color distortion while LLFlow generates a blurry image. Restormer and LLFormer fail to suppress the noise and contain over-/under-exposed regions. The bottom  row in Fig. \ref{Fig:qualitative} shows visual quality comparisons using an image from the LDIS dataset. Unlike the qualitative results observed from the top row, none of previous methods can achieve acceptable results on the image from LDIS, a challenging low-light dataset.  SOTA methods such as LLFlow and LLFormer fail to suppress the noise while Restormer introduces black spots and unnatural artifacts. Although our LDE-Net losses some texture details around shadow areas (e.g. under the chairs), it is capable of achieving satisfactory visual quality on lighting consistency, noise suppression, and structure detail. In summary, our proposed LDE-Net produces enhancement results with sharper details, comfortable contrast, and vivid colors, making it more satisfying and superior to other SOTA methods.

\begin{table}[t]
    \small
    \renewcommand{\arraystretch}{1.2}
    \begin{center}
    \resizebox{0.4\textwidth}{!}{
    \begin{tabular}{cccc|c|c}
        \toprule
          CAE & T-MSA & W-MSA & A-MSA  & PSNR$(\uparrow)$ & SSIM$(\uparrow)$ \\
        \hline
           $\checkmark$ & $\checkmark$ & & & 25.02 & 0.910 \\
        \hline
           & $\checkmark$ & & & 23.82 & 0.901 \\
        \hline
           & & $\checkmark$ & & 22.83 & 0.883 \\
       \hline
           & & & $\checkmark$  & 23.75 & 0.896 \\
      \bottomrule
    \end{tabular}
    }
    \end{center}
    \caption{We conduct ablation study on LOL-v1 dataset. PSNR and SSIM are reported}
    \vspace{-4mm}
    \label{tab:ablation study}
\end{table} 

\subsection{Nighttime UAV Tracking}
\label{sec: uav}

Benefiting from our latent disentanglement design as mentioned in Section \ref{sec: latent dis}, we further develop a light-weight enhancer ($\sim$ 0.05M parameters) for nighttime unmanned aerial vehicle (UAV) tracking tasks, as shown in Fig. \ref{Fig:light-enh}. We first use a $1 \times 1$ convolution to fuse the concatenation of disentangled Content and Illumination features. Then, a depth-wise separable $5 \times 5$ convolution is adopted to generate the deep Illumination feature. With another $1 \times 1$ convolution, the restored Illumination feature is output for further reconstruction. It is worth noting that  we only adopt the third partial of the enhancement loss in Eq. \ref{Eq: loss enh} with tracker backbone AlexNet \cite{bertinetto2016fully} as the task-specific loss to train the enhancer. 

\begin{table*}[t]
    \tiny
    \renewcommand{\arraystretch}{1.2}
    \begin{center}
    \resizebox{0.85\textwidth}{!}{
    \begin{tabular}{c|cccccc}
        \toprule
          Methods  & base-line \cite{redmon2018yolov3} & MBLLEN \cite{lv2018mbllen} & DeepLPF \cite{moran2020deeplpf} & Zero-DCE \cite{guo2020zero} & IAT \cite{cui2022you} & ours \\
        \hline
            mAP$(\uparrow)$ & 76.4 & 76.3& 76.3 & 76.9 & 77.2& 77.2 \\
    
        \hline
            FPS$(\uparrow) $ & 30.3 & 11.6  & 7.25  & 23.8 & 25.0 & 28.6 \\
 
      \bottomrule
    \end{tabular}
    }
    \end{center}
    \caption{Low-light detection results on ExDark \cite{loh2019getting} enhanced by different algorithms}
    \vspace{-4mm}
    \label{tab:low light object detection}
\end{table*}

\textbf{Experiment Settings.} We conduct nighttime UAV tracking on the DarkTrack2021 \cite{ye2022tracker} dataset to compare the preprocessing effects of our proposed enhancer for downstream vision understanding. The DarkTrack2021 dataset consists of 110 challenging sequences, 100,377 frames annotated with multiple object category bounding boxes (e.g. person, bus, car, truck, motor, dog, building, etc.), covering abundant scenarios of real-world UAV nighttime tracking tasks. To present the effectiveness of our enhancer against SCT \cite{ye2022tracker}, a specifically designed enhancer for nighttime UAV tracking, we choose 4 SOTA trackers, including SiamRPN++ \cite{li2019siamrpn++}, SiamAPN++ \cite{cao2021siamapn++}, DiMP18 and DiMP50 \cite{bhat2019learning}. Both our enhancer and the SCT enhancer are trained on LOL-v1 dataset. Following the one-pass evaluation (OPE) \cite{mueller2016benchmark}, we evaluate the tracking performance with two metrics, respectively precision and success rate. In addition, we also record the inference speed of trackers w./wo. the enhancer to illustrate the computational costs of plugging in a light-weight enhancer prior to SOTA trackers. The reported speed is evaluated on a machine with an
NVIDIA Titan Rtx GPU.

\textbf{Quantitative Results.} The success rate and precision scores are listed in Table \ref{tab:tracking results}. The results show that our light-weight enhancer significantly improve the nighttime tracking performance of all the trackers on DarkTrack2021 dataset. Compared with SCT, our enhancer also archieves a better success rate and precision incurring only low computational and memory costs.

\subsection{Low-light Object Detection}

\textbf{Experiment Settings.} We adopt the same pre-trained light-weight enhancer introduced in Section. \ref{sec: uav} for low-light object detection task. We conduct low-light object detection experiments on the ExDark \cite{loh2019getting} dataset to compare the preprocessing effects of different enhancement algorithms for high-level vision understanding. The ExDark dataset consists of 7363 under-exposed images annotated with 12 object category bounding boxes. 5890 images are selected for training while the left 1473 images are used for testing. YOLO-v3 \cite{redmon2018yolov3} is employed as the detector and trained from scratch. We use the average precision (AP) scores as the evaluation metric.

\textbf{Quantitative Results.} The detection metric mAP and evaluation speed is shown in Table \ref{tab:low light object detection}. Our model gains best results in both accuracy and speed compared to the baseline and other enhancement methods.

\subsection{Ablation Study}

In this section, we report two ablation studies we perform to demonstrate (i) the usefulness of our latent-space disentanglement approach (in terms of its performance and transferability), (ii) the effectiveness of various attention mechanisms and the proposed CAE module.

\textbf{Cross-Dataset Disentanglement.} We conduct a cross-dataset ablation study to compare our proposed latent disentanglement framework with the SOTA Retinex-based decomposition method Retinex-Net \cite{wei2018deep}. 
We evaluate our proposed latent disentanglement using paired low-/normal-light images $\mathbf{S}_l$ and $\mathbf{S}_t$. Based on the disentanglement objective, both $\mathbf{C}_l$ and $\mathbf{C}_t$ can be used to reconstruct a new image that is similar to the input image $\mathbf{S}_l / \mathbf{S}_t$ with the corresponding Illumination component $\mathbf{I}_l / \mathbf{I}_t$. Thus, we adopt PSNR metrics to calculate the difference between the reconstructed images and input images as shown in Table \ref{tab:cross-dataset dis} (e.g. $PSNR(M_{r}(\mathbf{C}_l,\mathbf{I}_l) , \mathbf{S}_l)$ represents the PSNR between $\mathbf{S}_l$ and the reconstructed image from $\mathbf{C}_l+\mathbf{I}_l$ ). Since the decomposition in Retinex-Net \cite{wei2018deep} share similar concept with our disentanglement approach, we use the same PSNR metrics to report their performance (For Retinex-Net, $\mathbf{C}$ denote the Reflectance image). 

We pre-train our disentanglement framework and the Decom-Net from Retinex-Net on LOL-v1 dataset and test on both LOL-v1 and LDIS datasets as shown in Table \ref{tab:cross-dataset dis}. The results on LOL-v1 dataset show that our method estimates much more consistent light-invariant components than Retinex-Net. Unlike Retinex-based decomposition, our disentanglement method doesn't introduce corruptions that need to be restored in the subsequent enhancement step. Further, the results on the LDIS dataset demonstrates that our method also outperforms the Retinex-Net in terms of the cross-dataset transferability and robustness under variant light conditions. Overall, all these results prove that our latent disentanglement design successfully separates the illumination component from a single image.

\textbf{Self-Attention Scheme.} We conduct an ablation to study the effectiveness of various self-attention schemes as shown in Table \ref{tab:ablation study}. For fair comparison, we first remove the CAE module and only implement 4-level  hierarchical encoder-decoder network for Illumination enhancement with different multi-head self-attention (MSA) mechanisms. We compared the transposed MSA (T-MSA) used in Restormer \cite{zamir2022restormer} with window-based MSA (W-MSA) proposed by Swin-Transformer and Axis-based MSA (A-MSA) proposed by LLFormer \cite{wang2023ultra}. The results show that T-MSA achieves the best performance.

\textbf{Adaptive Fusion.} The first row of Table \ref{tab:ablation study} shows the impact of adding the CAE module. The enhancement performance improves PSNR by 1.2 dB  with the CAE module which shows the usefulness of our CAE design.

%% file: Body/conclusions.tex
\section{Conclusions}

In this paper, we propose a novel transformer-based latent disentanglement framework to get better disentanglement results  for low-light image enhancement and other downstream low-light vision tasks. Then, we design a Content-Aware Embedding module (CAE), which explicitly improves the performance of Illumination enhancement by learning the correlation between the disentangled Content and Illumination features. Thirdly, we take the advantage of the proposed latent disentanglement framework to develop extremely light-weight enhancement network for downstream low-light vision tasks. Finally, extensive
experimental results 
demonstrate that our proposed Latent Disentangle-based Enhancement Network (LDE-Net) outperforms SOTA methods.